\documentclass[sigconf]{acmart}

\usepackage{booktabs} 

\setcopyright{rightsretained}

\acmDOI{}

\acmISBN{123-4567-24-567/08/06}

\acmConference[MLMH'18]{2018 KDD Workshop on Machine Learning for Medicine and Healthcare}{August 2018}{London, UK}
\acmYear{2018}
\copyrightyear{2018}
\acmArticle{4}
\acmPrice{15.00}

\begin{document}
\title{From Text to Topics in Healthcare Records: An Unsupervised Graph Partitioning Methodology}

\author{M. Tarik Altuncu}
\affiliation{%
  \institution{Dept. of Mathematics, Imperial College London}
  \streetaddress{South Kensington Campus}
}

\author{Erik Mayer}
\affiliation{%
  \institution{Dept. of Medicine, Imperial College London}
  \streetaddress{South Kensington Campus}
}

\author{Sophia N. Yaliraki}
\affiliation{%
  \institution{Dept. of Chemistry, Imperial College London}
  \streetaddress{South Kensington Campus}
}

\author{Mauricio Barahona}
\affiliation{%
  \institution{Dept. of Mathematics, Imperial College London}
  \streetaddress{South Kensington Campus}
}

\renewcommand{\shortauthors}{M.T. Altuncu et al}

\begin{abstract}
Electronic Healthcare Records contain large volumes of unstructured data, including extensive free text. Yet this source of detailed information often remains under-used  because of a lack of methodologies to extract interpretable content in a timely manner. Here we apply network-theoretical tools to analyse free text in Hospital Patient Incident reports from the National Health Service, to find clusters of documents with similar content in an unsupervised manner at different levels of resolution. We combine deep neural network paragraph vector text-embedding with multiscale Markov Stability community detection applied to a sparsified similarity graph of document vectors, and showcase the approach on incident reports from Imperial College Healthcare NHS Trust, London. The multiscale community structure reveals different levels of meaning in the topics of the dataset, as shown by descriptive terms extracted from the clusters of records. We also compare \textit{a posteriori} against hand-coded categories assigned by healthcare personnel, and show that our approach outperforms LDA-based models. Our content clusters exhibit good correspondence with two levels of hand-coded categories, yet they also provide further medical detail in certain areas and reveal complementary descriptors of incidents beyond the external classification taxonomy. 
\end{abstract}

%
%


\maketitle

\section{Introduction}
The vast amounts of data collected by healthcare providers in conjunction with modern data analytics techniques present a unique opportunity to improve the quality and safety of medical care, for patient benefit.  In the United Kingdom, the National Health Service (NHS) has a long history of documenting extensively the different aspects of healthcare provision. The NHS is currently in the process of increasing the availability of several databases, properly anonymised, with the aim to leverage advanced analytics to identify areas of improvement in its services. One such resource is 
the National Reporting and Learning System (NRLS), a central repository of patient safety incident reports from England and Wales collected since 2004, which now contains more than 13 million detailed records. The incidents are reported using a set of standardised categories and contain a wealth of organisational and spatiotemporal information (structured data) as well as, crucially, a substantial component of free text (unstructured data).
The incidents are wide ranging: from patient accidents to lost forms or referrals; from delays in admission and discharge to serious untoward incidents, such as retained foreign objects after operations. The review of such data provides critical insight into complex procedures in healthcare with an aim towards service improvement. 
 
Although statistical analyses are routinely performed on the structured components of the data, the free text component remains largely unused. Free text can be read manually but this task is time consuming, hence often ignored in practice.  Methods that provide automatic, content-based categorisation of incidents from the free text could help sidestep difficulties in assigning incident categories from \textit{a priori} pre-defined lists, reducing human error and burden, as well as offering a unique insight into the root cause analysis of incidents that could improve safety, quality of care and efficiency. 

Here, we showcase an algorithmic methodology that detects content-based groups of records in an unsupervised manner, based only on the free, unstructured textual description of the incidents.  To do so, we combine deep neural-network high-dimensional text-embedding algorithms (Doc2vec) with network-theoretical methods (multiscale Markov Stability (MS) community detection) applied to a sparsified geometric similarity graph of documents derived from text vector similarities. 

Traditional natural language processing tools have generally used bag-of-words representations of documents followed by statistical methods based on Latent Dirichlet Allocation (LDA) to cluster documents. More recent approaches have used deep neural network based language models, without a full multiscale graph analysis~\citep{Hashimoto2016TopicReviews}, while previous applications of network theory to text analysis
~\citep{PhysRevX.5.011007} were carried out at a single scale and used bag-of-word arrays lacking the power of neural network text embeddings. In contrast, multiscale community detection allows us to find groups of records with consistent content at different levels of resolution; hence the content categories emerge from the textual data, rather than fitting to pre-designed classifications. 

Our analysis starts by training a Doc2vec vector text embedding using the 13 million NRLS records. (We have also trained models to 1 and 2 million records and the results are similar.) Once the text embedding is obtained, we use it to produce document vectors for a subset of 3229 incident reports from St Mary's Hospital, London (Imperial College Healthcare NHS Trust) over three months in 2014. Our graph clustering method is then applied to cluster these records across different levels of resolution, revealing multiple levels of intrinsic structure in the topics of the dataset, as shown by the extraction of relevant word descriptors from the groups of records.

Upon reporting by the operator, the records had been independently hand-coded with up to 170 features per record, including a two-level manual classification of the incidents: 15 categories at Level 1; 95 sub-categories at Level 2. 
We carried out an \textit{a posteriori} comparison against the hand-coded categories assigned by the reporter. Several of our content-based clusters exhibit good correspondence with well-defined hand-coded categories at both levels; yet our results also provide further resolution in certain areas and a complementary characterisation of the incidents, not defined in the \textit{a priori} external classification.
We find that our methodology provides improved performance over LDA models, as quantified by the Uncertainty Coefficient against the hand-coded categories.

\section{Methodology}

{\bf{Text Pre-processing and Doc2Vec Model Training.}} 
The pre-processing of the raw text consisted of: lowering capital letters; tokenising sentences into words; stemming; and removing punctuation, stop-words and all numeric tokens. We then trained Doc2Vec models~\cite{doc2vec} using the PV-DBOW method from the Gensim~\cite{gensim} library. To ascertain the effect of the training of the Doc2Vec model on performance, we repeated the training with a broad range of parameters sets (window size, minimum count, subsampling). We carried Doc2Vec training both on a standard (generic, non-specialised) Wikipedia English corpus and on the full NRLS dataset (13+ million records with specialised language). 
Table~\ref{table:d2v} shows that, while the Wikipedia corpus is useful to train models when the parameters sets are weak, the NRLS dataset performs better for optimised parameter sets. Once optimised, the Doc2Vec model trained on the NRLS corpus is used to infer vectors for each of the $N=3229$ records in our analysis dataset.

\begin{table}[h]
\centering
\begin{tabular}{|l|l|l|l|l|}
\hline
\multicolumn{3}{|c|}{\textbf{Model Parameters}} & \multicolumn{2}{c|}{\textbf{Training Corpus}} \\ \hline
\textbf{\begin{tabular}[c]{@{}l@{}}Window\\  Size\end{tabular}} & \textbf{\begin{tabular}[c]{@{}l@{}}Minimum \\ Count\end{tabular}} & \textbf{Subsampling} & \textbf{Wikipedia} & \textbf{NRLS} \\ \hline
5 & 20 & 0.00001 & 465 & 379 \\ \hline
15 & 20 & 0.00001 & 424 & 387 \\ \hline
5 & 5 & 0.001 & 580 & 798 \\ \hline
5 & 20 & 0.001 & 587 & 809 \\ \hline
15 & 20 & 0.001 & 532 & 832 \\ \hline
15 & 5 & 0.001 & 531 & \bf{836} \\ \hline
\end{tabular}
\caption{The last two columns show the scores of Doc2Vec paragraph vector models trained using different hyper-parameter sets on different corpora. The scores are obtained by: (i) calculating centroids for the 15 hand-coded categories; 
(ii) selecting the 100 nearest reports for each centroid;
(iii) counting the number of incidents (out of 1500) correctly assigned to their centroid.}
\label{table:d2v}
\end{table}



{\bf{Graph Construction.}}   We constructed a normalised similarity matrix $\hat{S}$ based on the cosine similarity by: computing the matrix of cosine similarities between all pairs of records, $S_\text{cos}$; transforming it into a distance matrix $D_{cos} = 1-S_{cos}$; applying element-wise max norm to obtain $\hat{D}=\|D_{cos}\|_{max}$; and obtaining the normalized similarity matrix $\hat{S} = 1-\hat{D}$ which has values in $[0,1]$.
This (full) similarity matrix can be viewed as a completely connected, weighted graph. However, such a graph contains many edges with small weights (i.e., weak similarities) since, in high dimensional noisy datasets, even the least similar nodes present a non negligible degree of similarity. We thus apply a simple geometric sparsification to the normalized distance matrix $\hat{D}$  using the MST-kNN method\cite{mstknn}, a geometric heuristic that preserves the global connectivity of the graph while retaining the local geometry of the dataset. 
The MST-kNN graph is a weighted graph obtained by the union of the minimum spanning tree (MST) of $\hat{D}$, 
and adding edges connecting each node to its $k$ nearest nodes (kNN). 
We scanned $k={17, 13, 5, 1}$ for the graph construction and found that the MST-kNN graph with $k=13$ presents a reasonable balance between local and global structure in the dataset, and thus analyse it with the multi-scale graph partitioning framework. However, the results are robust to the choice of $k$ within a set of values. Note that the MST-kNN method avoids global similarity thresholding.

{\bf{Markov Stability Multiscale Graph Partitioning.}}
We apply Markov Stability (MS), a multiscale community detection method,  to the MST-kNN graph in order to detect clusters of documents with similar content 
at different levels of resolution. MS
is an unsupervised method that scans across all scales to detect robust and stable graph partitions using a continuous time diffusion process on the graph. The method does not need to choose {\it a priori} the number or type of relevant subgraphs, other than it retains the diffusive flow over the Markov time $t$. 
Hence $t$ acts as a resolution parameter revealing relevant partitions that persist over particular time scales in an unsupervised manner.  For more details see~\citep{pnasStability,Schaub2012ZoomingLens,LambiotteMarkovProcess,bacik_celegans}.

Briefly, the method optimises the MS function $r(t,H)$ over the space of graph partitions $H$ at each time $t$. MS is defined as the trace of the clustered autocovariance matrix of the diffusion process~\eqref{eq:MS}:
\begin{equation}
r(t,H) = \text{trace} \left[R(t,H)\right] = \text{trace} \left[H^T[\Pi e^{-t \mathcal{L}}-\pi\pi^T]H \right], \label{eq:MS}
\end{equation}
where $H$ is the membership matrix of the partition, $\mathcal{L}$ is the random walk Laplacian of the graph, $pi$ is the steady-state distribution of the process and $\Pi=\text{diag}(\pi)$. Our method searches for the partition $H^*(t)$ that maximises $r(t,H)$ at each Markov time. The partition $H^*(t)$ is formed by communities (subgraphs) that tend to preserve the flow within themselves over time $t$, since in that case the diagonal (off-diagonal) elements of $R(t,H)$ will be large (small). Although the maximization of~\eqref{eq:MS} is NP-Hard, there are optimisation methods that work well in practice. Here we use the Louvain Algorithm~\citep{louvain} which is efficient and known to give good results for benchmarks. We look to obtain robust partitions, i.e., partitions that are relevant across scales (i.e., consistently found over extended Markov time) and highly reproducible  (i.e., consistently found by the Louvain optimisation).  This is achieved by running the Louvain algorithm 500 times with different initialisations at each Markov time, picking the 50 with the highest MS value, and computing the variation of information $VI(t)$~\citep{Meila2007} of this ensemble. In addition, we also compute the variation of information between the optimised partitions found across time, $VI(t,t')$. 
Robust partitions are indicated by dips of $VI(t)$ and extended plateaux of $VI(t,t')$, indicating a partition that is robust to the optimisation and valid over extended scales~\citep{bacik_celegans,LambiotteMarkovProcess}. 



{\bf{Visualization and interpretation of results.} } 

\textit{Tracking membership through Sankey diagrams.} Sankey diagrams visualise the flow of node memberships across different partitions and categories. ~\nocite{sankey} 
We use Sankey diagrams with two different objectives: 
\textit{(i)} a multilayer Sankey diagram to represent the results of the multi resolution MS community detection across different scales (Fig.~\ref{fig:MSPlot});
\textit{(ii)} two-layer Sankey diagrams to indicate the correspondence between MS clusters and the hand-coded external categories at a given level of resolution (Fig.~\ref{fig:MS_17}).

\textit{Normalized contingency tables.} Normalised contingency tables allow us to compare the relative membership of content clusters in terms of the external categories. We plot contingency tables as heatmaps of a z-score (Fig.~\ref{fig:MS_17}). We score the quality of this correspondence using the uncertainty coefficient, an informational theoretical measure of similarity between groupings. 

\textit{Word clouds of increased intelligibility through lemmatisation.} To visualise the content of the document clusters, which can be understood as a type of \emph{topic detection}, we used Word Clouds weighted through lemmatisation as an intuitive way to summarise content and compare \textit{a posteriori} with hand-coded categories. Word clouds can also provide an aid for monitoring when used by practitioners. 

\section{Results}

\begin{figure}[h!]
\includegraphics[width=1\linewidth,angle=0]{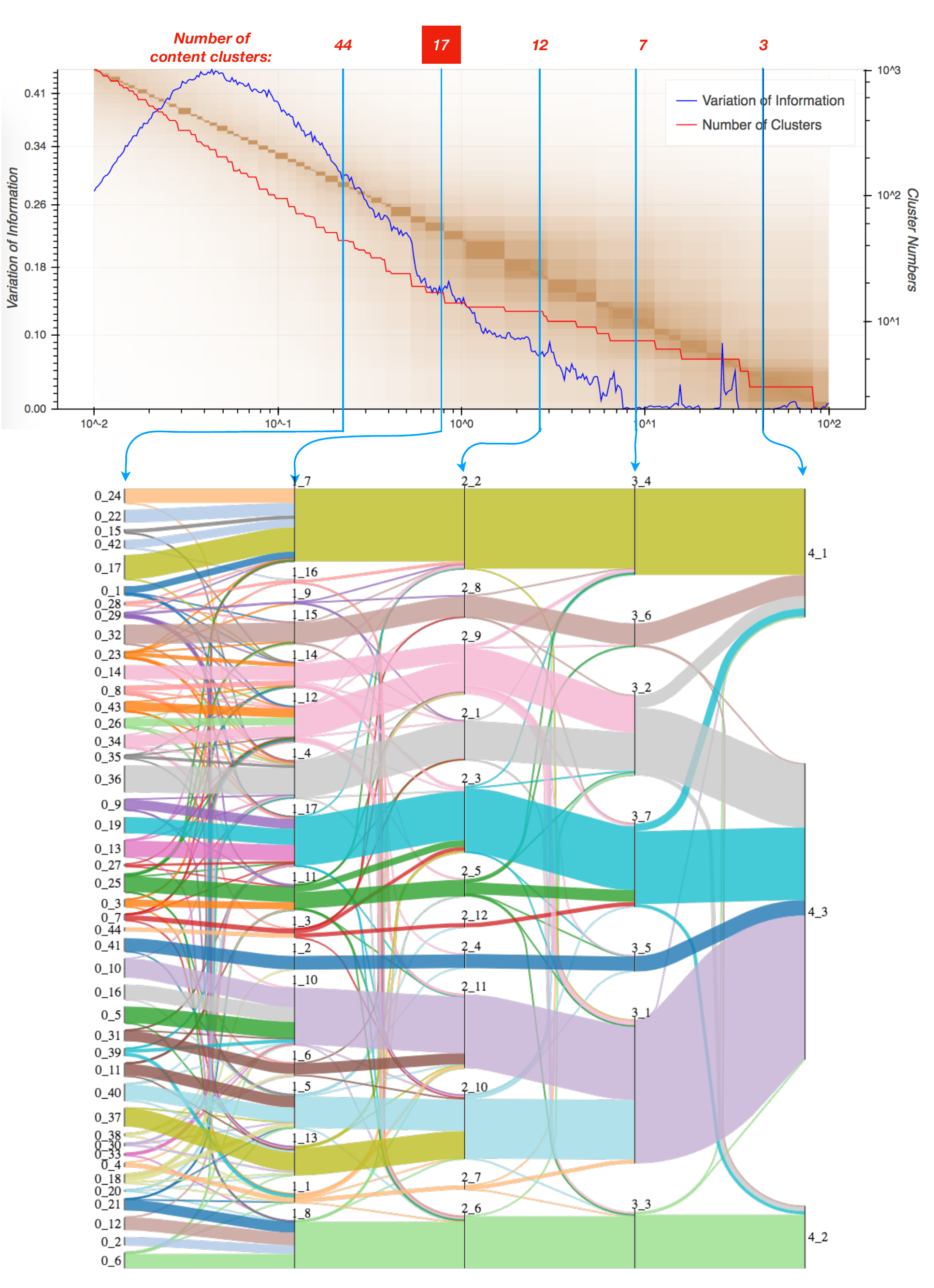}
\caption{
(Top) Markov Stability (MS) analysis across Markov time $t$: number of clusters of the optimised partitions (red line), $VI(t)$ for the ensemble of Louvain optimised solutions at each $t$ (blue line); $VI(t,t')$ between optimised partitions across Markov times (background colourmap). Relevant partitions (indicated by numbers and blue vertical lines) correspond to dips of $VI(t)$ and extended plateaux of $VI(t,t')$.  (Bottom) The Sankey diagram illustrates the quasi-hierarchical relationship of the communities of documents (indicated by numbers and colours) across levels of resolution.  
}
\label{fig:MSPlot}
\end{figure}

We applied full MS across an extended, finely sampled span of Markov times (0.01--100 in steps of 0.01) to the similarity MST-kNN graph of 3229 NRLS incident records. 
Figure~\ref{fig:MSPlot} presents a summary of this analysis including the number of clusters and the two metrics of variation of information  across all Markov times. The existence of several long plateaux in $VI(t,t')$ coupled to the existence of dips in the $VI(t)$ imply the presence of robust partitions at different levels of resolution. We choose several robust partitions, from finer to coarser, and examine their structure and content. Their relative node membership is shown with a multi-level Sankey diagram.

{\bf{Quasi-hierarchical structure: themes at different resolution levels. }}
The MS analysis in Figure~\ref{fig:MSPlot} reveals a rich multi-level structure of partitions, with a strong quasi-hierarchical organization. It is important to remark that, although the Markov time acts as a natural resolution parameter from finer to coarser partitions, our process of optimization does not impose such a hierarchical structure. Hence this organisation is intrinsic to the data and implies the existence of content communities which naturally integrate with each other as sub-themes of larger thematic categories. The detection of intrinsic scales within the graph at which robust partitions exist, thus allows us to obtain thematically-based clusters of records at different levels of resolution. 

\begin{figure}[h]
\includegraphics[width=1.05\linewidth,angle=0]{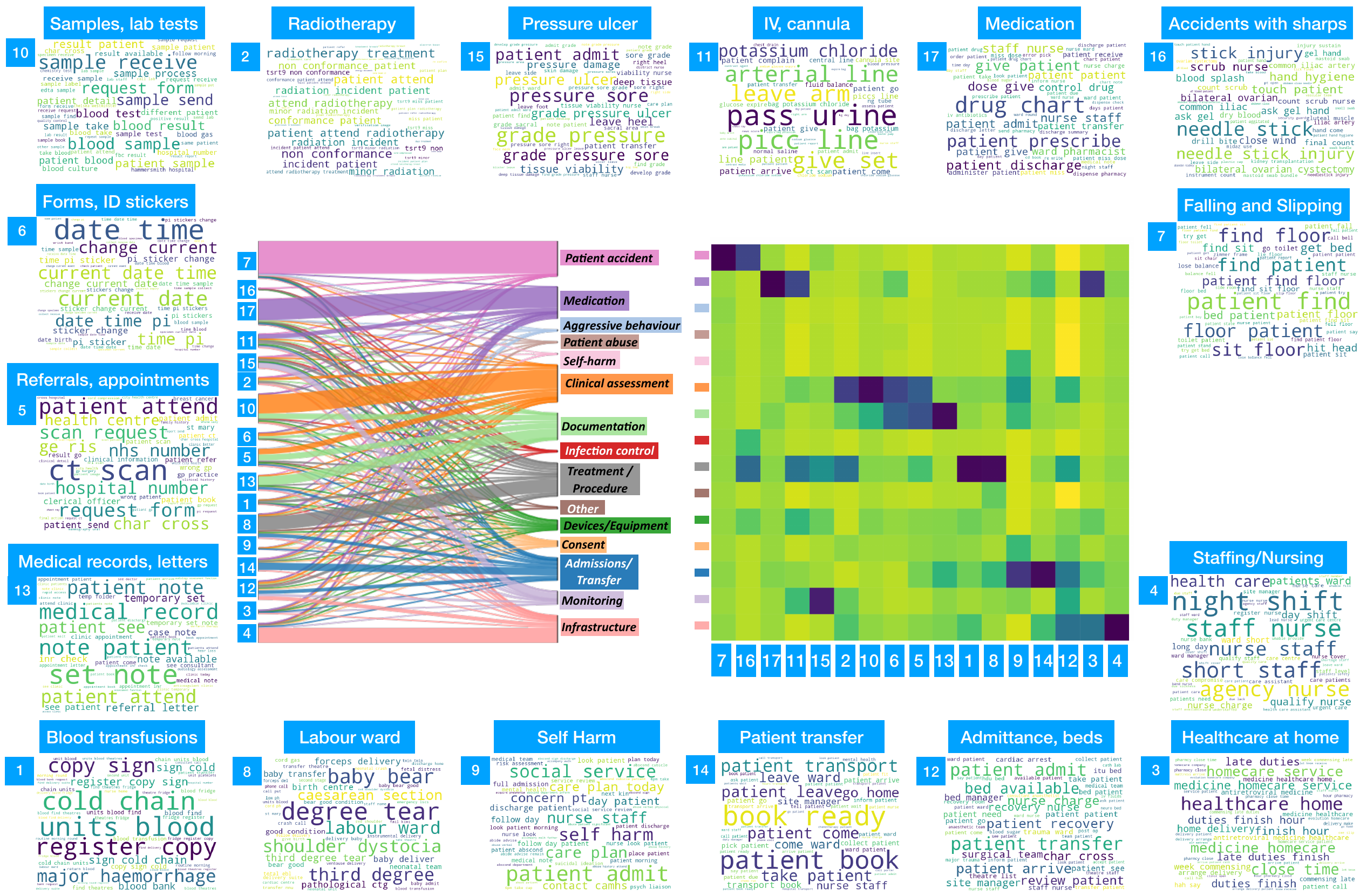}
\caption{
Summary of the 17-community MS partition. The content of each cluster is summarised with a word cloud (name tags given by us based only on the word cloud). The \textit{a posteriori} comparison to the 15 hand-coded categories (indicated by names and colours) is presented in two equivalent ways: a Sankey diagram showing the correspondence between categories and communities and the heatmap of a z-score contingency table.  
}
\label{fig:MS_17}
\end{figure}

{\bf{Interpretation of the MS communities: Word clouds and  \textit{a posteriori} comparison against hand-coded categories.}} To ascertain the relevance and relationship between the layers of MS clusters, we examined in detail the five levels of resolution in Figure~\ref{fig:MSPlot}. For each level, we prepared word clouds (lemmatised for increased intelligibility) and a Sankey diagram and contingency table linking content clusters with the hand-coded categories assigned by the operator. Note that this comparison was only done \textit{a posteriori}, i.e., the external categories were not used in our analysis, hence our approach is truly unsupervised. 
As an example, Figure~\ref{fig:MS_17} shows the 17-community partition with word clouds for all clusters and the comparison with the 15 hand-coded categories in Level 1.
%

\begin{figure}[h]
\includegraphics[width=.9 \linewidth,angle=0]{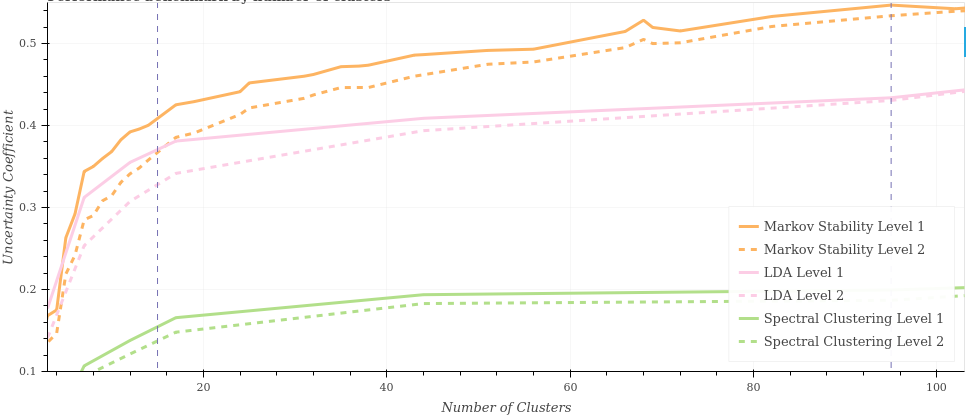}
\caption{The Uncertainty Coefficient measures the matching of LDA and MS clusterings over different scales against the hand-coded Level 1 and Level 2 categories. MS communities are consistently more coherent with the hand-coded categories than LDA and spectral clustering. The vertical dashed lines indicate the number of hand-coded categories for Level 1 and 2 (i.e., 15 and 95, respectively)
}
\label{fig:UC}
\end{figure}

{\bf{Comparison of the MS content clusters against other NLP methods.}}
We compared the MS document partitioning results against LDA models with a range of similar number of topics. LDA models are trained and inferred using the Gensim module on the same set of documents. We then quantify the match of the MS clusters with the hand-coded categories using the Uncertainty Coefficient \cite{uncoeff}
$$U(T|C)=\frac{H(T)-H(T|C)}{H(T)}=\frac{I(T;C)}{H(T)}$$ where $H(T)$ is the entropy of the hand-coded categories and $H(C)$ is the entropy of the clustering. The MS clusters show improved correspondence with both Level 1 and 2 categories as compared to LDA and spectral clustering methods (Fig.~\ref{fig:UC}).

\section{Discussion}
This work has applied a multiscale graph partitioning algorithm (MS) to determine topic clusters for a textual dataset of healthcare safety incident reports in an unsupervised manner at different levels of resolution. The method uses paragraph vectors to represent the records and obtains an ensuing similarity graph of documents constructed from their content. 
This method brings the advantage of multi-resolution algorithms capable of capturing clusters without imposing \textit{a priori} the number or structure of the clusters. Furthermore, it selects different levels of resolution of the clustering to suit the requirements of each task depending on the level of detail. 
The \textit{a posteriori} analysis against hand-categories showed that the method recovers meaningful categories and outperformed LDA at both categorisation levels. Furthermore, some of the MS content clusters capture topics of medical relevance, which provide complementary information to the external classifications.
The nuanced information and classifications extracted from free text analysis suggest a complementary axis to existing approaches to characterise patient safety incident reports, as 
the method allows for the discovery of emerging topics or classes of incidents directly from the data when such events do not fit the pre-assigned categories.

 
\bibliographystyle{ACM-Reference-Format}
\bibliography{appnetsci}


\begin{thebibliography}{13}


\ifx \showCODEN    \undefined \def \showCODEN     #1{\unskip}     \fi
\ifx \showDOI      \undefined \def \showDOI       #1{#1}\fi
\ifx \showISBNx    \undefined \def \showISBNx     #1{\unskip}     \fi
\ifx \showISBNxiii \undefined \def \showISBNxiii  #1{\unskip}     \fi
\ifx \showISSN     \undefined \def \showISSN      #1{\unskip}     \fi
\ifx \showLCCN     \undefined \def \showLCCN      #1{\unskip}     \fi
\ifx \shownote     \undefined \def \shownote      #1{#1}          \fi
\ifx \showarticletitle \undefined \def \showarticletitle #1{#1}   \fi
\ifx \showURL      \undefined \def \showURL       {\relax}        \fi
\providecommand\bibfield[2]{#2}
\providecommand\bibinfo[2]{#2}
\providecommand\natexlab[1]{#1}
\providecommand\showeprint[2][]{arXiv:#2}

\bibitem[\protect\citeauthoryear{Bacik, Schaub, Beguerisse-D{\'{i}}az, Billeh,
  and Barahona}{Bacik et~al\mbox{.}}{2016}]%
        {bacik_celegans}
\bibfield{author}{\bibinfo{person}{K~A Bacik}, \bibinfo{person}{M~T Schaub},
  \bibinfo{person}{M Beguerisse-D{\'{i}}az}, \bibinfo{person}{Y~N Billeh},
  {and} \bibinfo{person}{M Barahona}.} \bibinfo{year}{2016}\natexlab{}.
\newblock \showarticletitle{{Flow-Based Network Analysis of the Caenorhabditis
  elegans Connectome}}.
\newblock \bibinfo{journal}{\emph{PLOS Computational Biology}}
  \bibinfo{volume}{12}, \bibinfo{number}{8} (\bibinfo{year}{2016}),
  \bibinfo{pages}{1--27}.
\newblock


\bibitem[\protect\citeauthoryear{Blondel, Guillaume, Lambiotte, and
  Lefebvre}{Blondel et~al\mbox{.}}{2008}]%
        {louvain}
\bibfield{author}{\bibinfo{person}{V~D Blondel}, \bibinfo{person}{J-L
  Guillaume}, \bibinfo{person}{R Lambiotte}, {and} \bibinfo{person}{E
  Lefebvre}.} \bibinfo{year}{2008}\natexlab{}.
\newblock \showarticletitle{{Fast unfolding of communities in large networks}}.
\newblock \bibinfo{journal}{\emph{Journal of Statistical Mechanics: Theory and
  Experiment}} \bibinfo{volume}{2008}, \bibinfo{number}{10}
  (\bibinfo{year}{2008}), \bibinfo{pages}{P10008}.
\newblock
\showISBNx{1742-5468}


\bibitem[\protect\citeauthoryear{Delvenne, Yaliraki, and Barahona}{Delvenne
  et~al\mbox{.}}{2010}]%
        {pnasStability}
\bibfield{author}{\bibinfo{person}{J-C Delvenne}, \bibinfo{person}{S~N
  Yaliraki}, {and} \bibinfo{person}{M Barahona}.}
  \bibinfo{year}{2010}\natexlab{}.
\newblock \showarticletitle{{Stability of graph communities across time
  scales.}}
\newblock \bibinfo{journal}{\emph{Proceedings of the National Academy of
  Sciences of the United States of America}} \bibinfo{volume}{107},
  \bibinfo{number}{29} (\bibinfo{date}{7} \bibinfo{year}{2010}),
  \bibinfo{pages}{12755--60}.
\newblock


\bibitem[\protect\citeauthoryear{Hashimoto, Kontonatsios, Miwa, and
  Ananiadou}{Hashimoto et~al\mbox{.}}{2016}]%
        {Hashimoto2016TopicReviews}
\bibfield{author}{\bibinfo{person}{K Hashimoto}, \bibinfo{person}{G
  Kontonatsios}, \bibinfo{person}{M Miwa}, {and} \bibinfo{person}{S
  Ananiadou}.} \bibinfo{year}{2016}\natexlab{}.
\newblock \showarticletitle{{Topic detection using paragraph vectors to support
  active learning in systematic reviews}}.
\newblock \bibinfo{journal}{\emph{Journal of Biomedical Informatics}}
  \bibinfo{volume}{62} (\bibinfo{date}{8} \bibinfo{year}{2016}),
  \bibinfo{pages}{59--65}.
\newblock


\bibitem[\protect\citeauthoryear{Lambiotte, Delvenne, and Barahona}{Lambiotte
  et~al\mbox{.}}{2014}]%
        {LambiotteMarkovProcess}
\bibfield{author}{\bibinfo{person}{R Lambiotte}, \bibinfo{person}{J~C
  Delvenne}, {and} \bibinfo{person}{M Barahona}.}
  \bibinfo{year}{2014}\natexlab{}.
\newblock \showarticletitle{{Random Walks, Markov Processes and the Multiscale
  Modular Organization of Complex Networks}}.
\newblock \bibinfo{journal}{\emph{IEEE Transactions on Network Science and
  Engineering}} \bibinfo{volume}{1}, \bibinfo{number}{2} (\bibinfo{date}{7}
  \bibinfo{year}{2014}), \bibinfo{pages}{76--90}.
\newblock
\showISSN{2327-4697}
\urldef\tempurl%
\url{https://doi.org/10.1109/TNSE.2015.2391998}
\showDOI{\tempurl}


\bibitem[\protect\citeauthoryear{Lancichinetti, Sirer, Wang, Acuna,
  K{\"{o}}rding, and Amaral}{Lancichinetti et~al\mbox{.}}{[n. d.]}]%
        {PhysRevX.5.011007}
\bibfield{author}{\bibinfo{person}{A Lancichinetti}, \bibinfo{person}{M~I
  Sirer}, \bibinfo{person}{J~X Wang}, \bibinfo{person}{D Acuna},
  \bibinfo{person}{K K{\"{o}}rding}, {and} \bibinfo{person}{L~N Amaral}.}
  \bibinfo{year}{[n. d.]}\natexlab{}.
\newblock \showarticletitle{{High-Reproducibility and High-Accuracy Method for
  Automated Topic Classification}}.
\newblock \bibinfo{journal}{\emph{Phys. Rev. X}} \bibinfo{number}{1}
  (\bibinfo{date}{jan} \bibinfo{year}{[n. d.]}), \bibinfo{pages}{11007}.
\newblock
\urldef\tempurl%
\url{https://doi.org/10.1103/PhysRevX.5.011007}
\showDOI{\tempurl}


\bibitem[\protect\citeauthoryear{Le and Mikolov}{Le and Mikolov}{2014}]%
        {doc2vec}
\bibfield{author}{\bibinfo{person}{Q~Qv Le} {and} \bibinfo{person}{T Mikolov}.}
  \bibinfo{year}{2014}\natexlab{}.
\newblock \showarticletitle{{Distributed Representations of Sentences and
  Documents}}.
\newblock \bibinfo{journal}{\emph{International Conference on Machine Learning
  - ICML 2014}}  \bibinfo{volume}{32} (\bibinfo{year}{2014}),
  \bibinfo{pages}{1188–1196}.
\newblock
\showISBNx{9781634393973}
\showISSN{10495258}
\urldef\tempurl%
\url{https://doi.org/10.1145/2740908.2742760}
\showDOI{\tempurl}


\bibitem[\protect\citeauthoryear{Lupton and {Leonardo}}{Lupton and
  {Leonardo}}{2017}]%
        {sankey}
\bibfield{author}{\bibinfo{person}{R Lupton} {and}
  \bibinfo{person}{{Leonardo}}.} \bibinfo{year}{2017}\natexlab{}.
\newblock \bibinfo{title}{{ricklupton/sankeyview: v1.1.7}}.
\newblock
\newblock
\urldef\tempurl%
\url{https://doi.org/10.5281/zenodo.1098904}
\showDOI{\tempurl}


\bibitem[\protect\citeauthoryear{Meil{\u{a}}}{Meil{\u{a}}}{2007}]%
        {Meila2007}
\bibfield{author}{\bibinfo{person}{Marina Meil{\u{a}}}.}
  \bibinfo{year}{2007}\natexlab{}.
\newblock \showarticletitle{{Comparing clusterings—an information based
  distance}}.
\newblock \bibinfo{journal}{\emph{Journal of Multivariate Analysis}}
  \bibinfo{volume}{98}, \bibinfo{number}{5} (\bibinfo{date}{5}
  \bibinfo{year}{2007}), \bibinfo{pages}{873--895}.
\newblock


\bibitem[\protect\citeauthoryear{Rehurek and Sojka}{Rehurek and Sojka}{2010}]%
        {gensim}
\bibfield{author}{\bibinfo{person}{R Rehurek} {and} \bibinfo{person}{P Sojka}.}
  \bibinfo{year}{2010}\natexlab{}.
\newblock \showarticletitle{{Software Framework for Topic Modelling with Large
  Corpora}}. In \bibinfo{booktitle}{\emph{Proceedings of the LREC 2010 Workshop
  on New Challenges for NLP Frameworks}}. \bibinfo{publisher}{ELRA},
  \bibinfo{address}{Valletta, Malta}, \bibinfo{pages}{45--50}.
\newblock


\bibitem[\protect\citeauthoryear{Schaub, Delvenne, Yaliraki, and
  Barahona}{Schaub et~al\mbox{.}}{2012}]%
        {Schaub2012ZoomingLens}
\bibfield{author}{\bibinfo{person}{M~T Schaub}, \bibinfo{person}{J~C Delvenne},
  \bibinfo{person}{S~N. Yaliraki}, {and} \bibinfo{person}{M Barahona}.}
  \bibinfo{year}{2012}\natexlab{}.
\newblock \showarticletitle{{Markov dynamics as a zooming lens for multiscale
  community detection: Non clique-like communities and the field-of-view
  limit}}.
\newblock \bibinfo{journal}{\emph{PLoS ONE}} (\bibinfo{year}{2012}).
\newblock
\showISSN{19326203}
\urldef\tempurl%
\url{https://doi.org/10.1371/journal.pone.0032210}
\showDOI{\tempurl}


\bibitem[\protect\citeauthoryear{Veenstra, Cooper, and Phelps}{Veenstra
  et~al\mbox{.}}{2017}]%
        {mstknn}
\bibfield{author}{\bibinfo{person}{P Veenstra}, \bibinfo{person}{C Cooper},
  {and} \bibinfo{person}{S Phelps}.} \bibinfo{year}{2017}\natexlab{}.
\newblock \showarticletitle{{Spectral clustering using the kNN-MST similarity
  graph}}.
\newblock In \bibinfo{booktitle}{\emph{2016 8th Computer Science and Electronic
  Engineering Conference, CEEC 2016 - Conference Proceedings}}.
  \bibinfo{publisher}{Institute of Electrical and Electronics Engineers Inc.},
  \bibinfo{pages}{222--227}.
\newblock
\showISBNx{9781509020508}
\urldef\tempurl%
\url{https://doi.org/10.1109/CEEC.2016.7835917}
\showDOI{\tempurl}


\bibitem[\protect\citeauthoryear{White, Steingold, and Fournelle}{White
  et~al\mbox{.}}{2004}]%
        {uncoeff}
\bibfield{author}{\bibinfo{person}{JV White}, \bibinfo{person}{Sam Steingold},
  {and} \bibinfo{person}{CG Fournelle}.} \bibinfo{year}{2004}\natexlab{}.
\newblock \showarticletitle{Performance metrics for group-detection
  algorithms}.
\newblock \bibinfo{journal}{\emph{Proceedings of Interface 2004}}
  (\bibinfo{year}{2004}).
\newblock


\end{thebibliography}

\end{document}